\documentclass[sigconf,9pt, nonacm]{acmart}

\usepackage{booktabs}
\usepackage{siunitx}
\usepackage{numprint}

\usepackage{tikz}
\usepackage{amsmath}
\usepackage{bbm}
\usepackage{algpseudocode}

\usepackage{colortbl}
\usepackage{eucal}
\usepackage{bm}
\usepackage{subfigure}
\usepackage{booktabs}
\usepackage{multirow}
\usepackage{enumitem}
\usepackage{url}
\usepackage{xspace}
\usepackage{amsfonts}
\usepackage{booktabs}
\usepackage{graphicx}
\usepackage{svg}

\usepackage[table,xcdraw]{xcolor}
\definecolor{deepsea}{RGB}{135, 206, 235}

\usepackage{algorithm}
\usepackage{pifont}

\usepackage{makecell} 

\newcolumntype{C}[1]{>{\centering\arraybackslash}p{#1}}
\newcolumntype{L}[1]{>{\raggedright\arraybackslash}p{#1}}

\setcopyright{acmlicensed}
\copyrightyear{2018}
\acmYear{2018}
\acmDOI{XXXXXXX.XXXXXXX}
\acmConference[Conference acronym 'XX]{Make sure to enter the correct
  conference title from your rights confirmation email}{June 03--05,
  2018}{Woodstock, NY}
\acmISBN{978-1-4503-XXXX-X/2018/06}

\def\blueNode{{blue node}\xspace} 
\def\grayNode{{gray node}\xspace} 
\def\blueNodes{{blue nodes}\xspace} 
\def\grayNodes{{gray nodes}\xspace} 

\def\moduleOne{{Packet Grouping}\xspace} 
\def\moduleTwo{{Tracker}\xspace}

\def\ourSystem{{GraySense}\xspace} 

\begin{document}

\title{Tracking without Seeing: Geospatial Inference using Encrypted Traffic from Distributed Nodes}

\author{%
  Sadik~Yagiz~Yetim$^{1*}$, 
  Gaofeng~Dong$^{1*}$, 
  Isaac-Neil~Zanoria$^{1*}$, 
  Ronit~Barman$^{1}$, 
  Maggie~Wigness$^{2}$, 
  Tarek~Abdelzaher$^{3}$, 
  Mani~Srivastava$^{1}$, 
  Suhas~Diggavi$^{1}$%
  \\\normalsize
  \{yagizyetim, gfdong, zanoria, ronitbarman\}@ucla.edu,  maggie.b.wigness.civ@army.mil, zaher@illinois.edu, \{mbs, suhas\}@ucla.edu 
  \normalsize
  $^{1}$University of California, Los Angeles \quad
  $^{2}$DEVCOM Army Research Laboratory \quad
  $^{3}$University of Illinois at Urbana-Champaign \quad
}
\thanks{$^{*}$Equal contribution.}

\begin{abstract}

Accurate observation of dynamic environments traditionally relies on synthesizing raw, signal-level information from multiple distributed sensors. This work investigates an alternative approach: performing geospatial inference using only encrypted packet-level information, without access to the raw sensory data itself. We further explore how this indirect information can be fused with directly available sensory data to extend overall inference capabilities.
In this paper, we introduce \ourSystem, a learning-based framework that performs geospatial object tracking by analyzing indirect information from cameras with inaccessible streams. This information consists of encrypted wireless video transmission traffic, such as network-level packet sizes.
\ourSystem leverages the inherent relationship between scene dynamics and the transmitted packet sizes of the video streams to infer object motion. 
The framework consists of two stages: (1) a \moduleOne module that identifies frame boundaries and estimates frame sizes from encrypted network traffic, and (2) a \moduleTwo module, based on a Transformer encoder with a recurrent state, which fuses these indirect, packet-based inputs with optional direct, camera-based, sensory inputs to estimate the object's position.
Extensive experiments, conducted with realistic videos from the CARLA simulator and emulated networks under varying, imperfect conditions, show that \ourSystem achieves a high tracking accuracy of $2.33$ meters error (Euclidean distance) \textit{without} raw signal access. This error is reasonable relative to the smallest tracked object’s dimensions ($4.61 \text{m} \times 1.93 \text{m}$), enabling meaningful trajectory estimation.
In summary, we demonstrate a method that performs geospatial inference using only encrypted traffic, without access to raw signals. To the best of our knowledge, such capability has not been previously observed, and it expands the use of latent signals available for sensing.

\end{abstract}

\maketitle
\pagestyle{plain}

\section{Introduction}\label{sec_introduction}

\begin{figure}[!htp]
    \centering
    \includegraphics[width=0.7\linewidth]{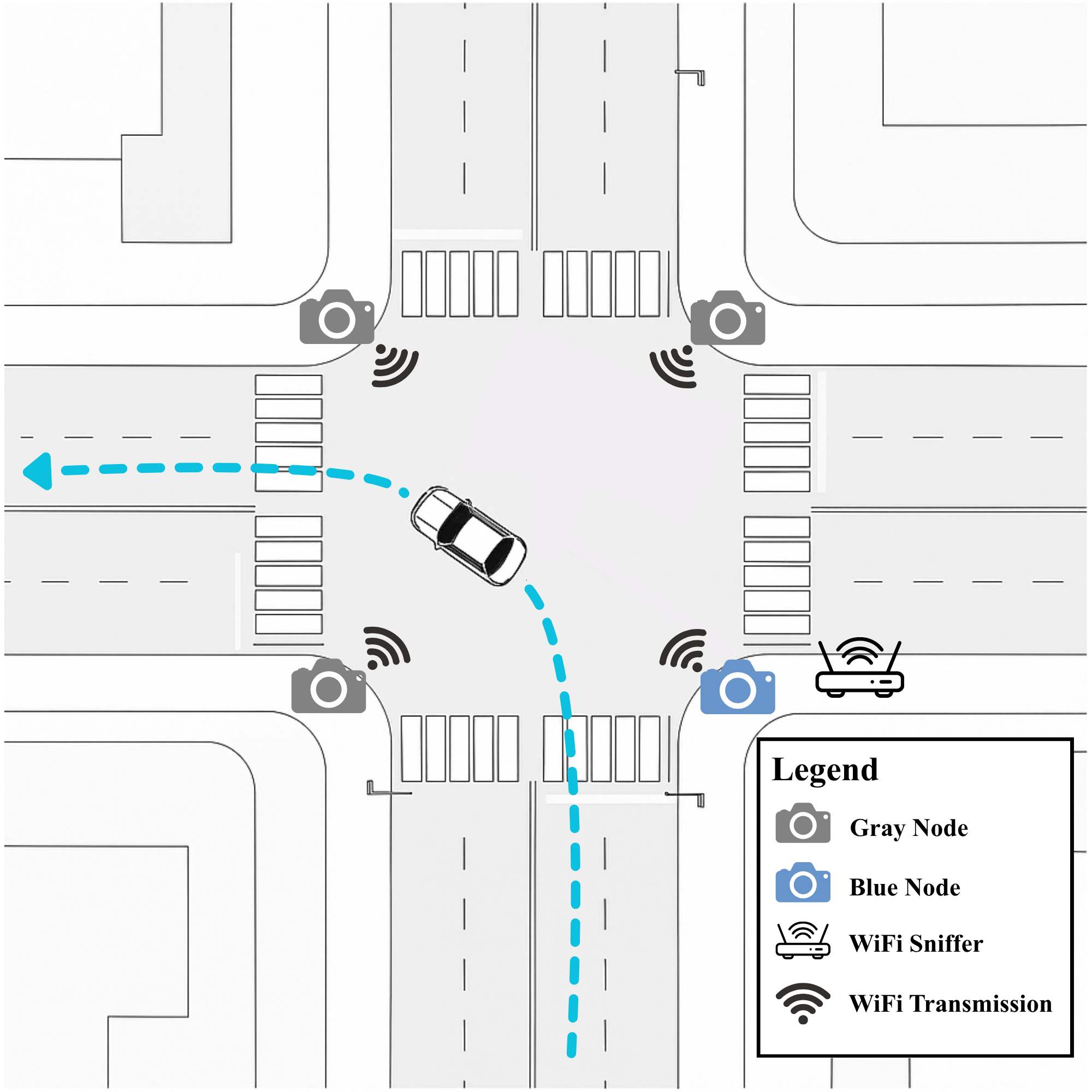}
    \caption{Application scenario of \ourSystem. 
    The scene illustrates a distributed sensing environment consisting of accessible cameras (\blueNodes{}) and inaccessible cameras (\grayNodes{}) whose video streams are encrypted. 
    While \blueNodes{} provide direct visual input, \grayNodes{} contribute only encrypted network traffic, which indirectly reflects scene dynamics through variations in packet sizes. 
    }
    \label{fig:scenario}
    \vspace{-0.15in}
\end{figure}

A comprehensive understanding of dynamic environments increasingly depends on integrating information from multiple distributed sensors, whose complementary viewpoints and spatial coverage together provide richer and more resilient perception of the scene~\cite{samplawski2023heteroskedastic,wang2022detr3d, han2024exploring}.
However, in practice, direct access to raw sensor data can be limited. Ownership boundaries, privacy policies, and encryption protocols restrict systems to operate with only a subset of all available sensors~\cite{ouyang2025mmbind,liu2020handling}. Meanwhile, many nearby sensors remain physically co-located yet logically isolated, broadcasting encrypted network traffic that indirectly reflects their sensing activity~\cite{huang2023rethinking,rasool2025invisible,mari2021looking}. These inaccessible nodes form a latent layer of the environment’s perceptual network - one that cannot be directly observed, yet whose transmission dynamics encode meaningful information about the underlying scene. 
Unlocking this information could extend opportunistic perception,
where a system observes a scene not just by “seeing” directly, but by listening to the data flows of surrounding sensors.
In this work, we use the term “seeing” to refer broadly to having access to raw, signal-level sensory measurements. Thus, “tracking without seeing” denotes the capability to perform geospatial inference without access to raw sensor signal.

This asymmetry between accessible and inaccessible sensors motivates our exploration of "indirect" sensing. We categorize sensors into two classes based on their accessibility: \textit{\textbf{\blueNodes}}, which denote sensors directly accessible to the system (e.g., cameras with available video streams), and \textit{\textbf{\grayNodes}}, which represent indirect sensors whose measurements cannot be accessed but encrypted transmission traffic can be passively observed. 

In this paper, we exclusively use \textbf{cameras as sensors}, as illustrated in Figure~\ref{fig:scenario}. While \blueNodes provide conventional visual information, \grayNodes contribute only network-level information such as packet sizes. Despite this constraint, the temporal patterns of \grayNode traffic can still provide rich information regarding the innovations in the underlying scene depending on the transmission protocol. Popular video codecs, such as H.264~\cite{wiegand2003overview}, primarily encode the differences between consecutive frames. This process results in significant frame-size variations that are preserved even after encryption, causing the observed traffic to inherently carry structured information on scene dynamics, which we exploit through a learning-based framework to do geospatial inference.\footnote{We believe the overall idea of using structure in packets induced by compression can be applied to other formats as well, which are discussed in Section~\ref{subsec:back_gop}.}

Recent studies have shown that encrypted video streams can provide information on the dynamics of the monitored environment. 
For instance, Li et al.~\cite{li2016side} demonstrated that side-channel information in encrypted surveillance video traffic can be exploited to infer coarse activity patterns such as the presence or movement of people within a scene. Similarly, Huang et al.~\cite{huang2023rethinking} highlighted privacy risks from wireless surveillance cameras by showing that encrypted traffic features can reveal basic human activities. Mari et al.~\cite{mari2021looking} went beyond activity detection to infer rough walking directions from encrypted video-surveillance traffic. Rasool et al.~\cite{rasool2025invisible} further developed non-machine-learning methods for detecting motion and person presence. 
However, these approaches do not analyze how frame-level dynamics relate to the underlying scene geometry, and typically rely on traffic from a single \grayNode. Furthermore, they treat the problem as a classification or detection task without explicitly modeling or reconstructing any spatial information from the scene.
Building on these prior work on single-object sensing, \ourSystem prioritizes single-target tracking to match monitoring requirements in high-stakes security perimeters, where a single intruder constitutes a critical state transition and maintaining robust position estimation is the primary objective.
This aligns with single-threat validation protocols used in sterile-zone surveillance \cite{sandia_sttem, nist_ir_7972}. A similar framing appears in search and rescue, where the goal is often to locate a specific target \cite{uav_sar_tracking}. Extending \ourSystem to the multi-object setting is discussed in Section~\ref{sec_dis}.

In this work, we systematically investigate the relationship between scene dynamics and encrypted traffic patterns, and propose the \ourSystem framework for geospatial object tracking using encrypted WiFi camera traffic. 
By leveraging multiple distributed sensors, \blueNodes\ and \grayNodes,  \textbf{we demonstrate that it is possible not only to detect motion but to continuously track an object's position from the encrypted network traffic}, revealing a previously unexplored capability, 
=which offers several compelling advantages. By relying solely on packet-size information in video transmissions, \ourSystem enables object \textbf{tracking without access to raw camera streams}, providing an alternative to conventional vision-based tracking. Moreover, when combined with available \blueNodes, information from surrounding \grayNodes can extend the effective sensing range and maintain target continuity even after the object leaves the field of view of accessible cameras. Finally, since it operates on lightweight network metadata rather than raw image or video data, \ourSystem brings significant reductions in computational and communication complexity, making it well-suited for scalable deployment across distributed or resource-constrained sensing systems.

Although promising, \ourSystem faces several challenges.
First, extracting useful frame-level information, such as frame boundaries and relative frame sizes, from encrypted network traffic is nontrivial, especially under network imperfections such as bandwidth limit, variable delay, and jitter. These effects obscure temporal regularities and make frame segmentation unreliable.
Second, while packet-level traffic from a single gray node encodes aggregate temporal patterns that correlate with scene dynamics, it conveys only a total measure of change, without any explicit spatial context of where this change occurs. In other words, packet sizes provide a compressed, viewpoint-specific signal that integrates motion over the entire field of view, making spatial localization highly challenging. This limitation introduces significant ambiguity, as multiple scene configurations can produce similar packet-size sequences.
Third, the absence of existing datasets that couple ground-truth trajectories, video content, and realistic network traces hinders the development and evaluation of learning-based approaches.

To address these challenges, we introduce \ourSystem with following advances. 
First, we develop a \textbf{\moduleOne module} used for inferring frame boundaries from raw encrypted packet streams, reconstructing frame-level structure even under noisy network conditions.
Second, we develop a \textbf{\moduleTwo module} built on a Transformer encoder \cite{you_just_want} with a recurrent state, which fuses information from multiple distributed \grayNodes and optional \blueNodes. By leveraging complementary viewpoints across these sensors, the module jointly learns from multi-view encrypted streams to recover spatial structure from non-spatial measurements, enabling accurate trajectory estimation.
Third, to enable systematic training and validation, we construct a suite of \textbf{realistic synthetic datasets using the CARLA simulator coupled with controlled network emulation}, capturing diverse traffic conditions, trajectories, and environmental variations.

Extensive experiments demonstrate that \ourSystem achieves $2.33$m tracking error using only the encrypted traffic from four \grayNodes, under noisy network conditions. 
These results demonstrate that accurate object tracking is achievable using only encrypted network traffic, without accessing any sensor directly. This work validates the \textbf{concept of tracking without seeing} through \textbf{\ourSystem}, which extends the sensing capabilities of \blueNodes by leveraging \grayNodes and opens new avenues for future research.

Our key contributions are summarized as follows: 
\begin{itemize}[label=\textbullet, leftmargin=1em, topsep=0em]

    \item We develop a novel framework for geospatial object tracking using encrypted traffic from \grayNodes and optional video streams from \blueNodes within distributed sensing systems, and demonstrate its underlying feasibility through geometric analysis.
        
    \item We propose a two-stage learning framework that performs tracking by combining a \moduleOne module with a Transformer-based \moduleTwo network, designing a customized loss function and a recurrent-state design to enable consistent tracking.
    
    \item We construct a comprehensive suite of realistic synthetic datasets covering diverse scenarios and network conditions, which will be publicly released upon acceptance to foster future research\footnote{https://github.com/nesl/graysense}. Extensive experiments validate the effectiveness of \ourSystem, demonstrating accurate and robust tracking performance under a wide range of sensing and network configurations.    

\end{itemize}

\section{Related Work}\label{sec_relatedwork}

This work connects to several research directions, including wireless side-channel analysis, human activity inference from WiFi traffic, and signal-based camera eavesdropping. We briefly discuss their relevance and contrast them with our approach.

\subsection{Locating Hidden Cameras via WiFi Traffic}
Recent efforts have leveraged wireless side channels to detect or localize hidden cameras through analysis of their WiFi transmissions. Systems such as SnoopDog~\cite{singh2021always}, Lumos~\cite{sharma2022lumos}, and LocCams~\cite{gu2024loccams} identify the presence and position of surveillance devices by introducing controlled environmental motion and correlating the resulting traffic patterns with known motions.
In contrast, our work assumes that camera parameters are already known and focuses on the inverse problem - inferring object motion and trajectory from encrypted traffic emitted by known cameras. Rather than detecting cameras, we use them as sources of motion-related information, turning their encrypted transmissions into an indirect sensing modality.

\subsection{Inferring Human Motion from WiFi Traffic}
A parallel line of research explores using WiFi traffic itself to infer human activities or coarse spatial regions in smart-home or indoor settings.
Li et al.~\cite{li2016side} and Huang et al.~\cite{huang2023rethinking} demonstrated that encrypted video streams leak information correlated with physical motion, enabling detection of human activities. Rasool et al.~\cite{rasool2025invisible} proposed a non-machine-learning method for live-streaming and motion detection through encrypted traffic analysis, while Mari et al.~\cite{mari2021looking} further inferred walking directions from video-surveillance data.
These studies reveal that encrypted network traffic contains latent motion cues but are limited to classification or detection tasks.
In contrast, our framework performs continuous geospatial tracking, reconstructing object trajectories rather than categorizing activities, and extends the analysis to multi-node, distributed settings with both accessible (blue) and inaccessible (gray) sensors.

\subsection{Tracking with Compressed Domain Information}
Prior studies have explored object detection and tracking using information from the compressed video domain rather than raw pixels. MVmed~\cite{9248145} combines a standard pixel-domain detector (e.g., Faster R-CNN) that runs intermittently on I-frames with a high-speed compressed-domain tracker, predicting object locations between full detections by averaging motion vectors within the last known bounding box. 
A complementary line of work, Moustafa et al.~\cite{moustafa2021deep} directly feed sparse residual frames into a neural network, showing that residuals alone can serve as lightweight and privacy-preserving tracking features, though they still require access to decoded motion information rather than encrypted data. 
More recently, Tian et al.~\cite{10262343} proposed a secure deep learning framework for moving object detection in compressed video using Encrypted Domain Motion Information (EDMI). Their method operates without full decryption or decompression by designing three motion feature maps derived from intentionally unencrypted coding features such as partition patterns and the number of coding bits at the block level. These features provide a coarse spatial representation of motion, revealing which 16×16 blocks are complex or frequently subdivided, thus enabling low-resolution motion inference while preserving partial encryption. In contrast, our framework does not assume any selectively unencrypted features or access to codec internals, instead inferring scene dynamics solely from encrypted packet-level traffic.

\subsection{Recovering Images from Electromagnetic Side Channels}
Orthogonal to network-based approaches, EM Eye~\cite{long2024eye} demonstrates that electromagnetic (EM) emissions from embedded cameras can be exploited to reconstruct images of the observed scene. While impressive in fidelity, this method requires close proximity (typically within meter level) and dense sensor placement due to the rapid signal attenuation of EM leakage.
Our approach instead operates over standard wireless network channels, offering a longer effective range. \ourSystem aims to recover motion-level information rather than pixel-level imagery, providing a scalable alternative to physical side-channel imaging.

\subsection{Tracking with Non-visual Modalities} 
A rich body of research has investigated non-visual modalities to augment visual sensing and tracking, such as acoustic~\cite{schulz2021hearing,hao2024acoustic}, ToF~\cite{kadambi2016occluded}, and radar systems~\cite{rabaste2017around, scheiner2020seeing}. Distinct from these efforts, \ourSystem\ enables inference even without access to raw signals, using only encrypted traffic from \grayNodes.

\section{Background and Problem Formulation}\label{sec_back}

\begin{figure}
    \centering
    \includegraphics[width=0.8\linewidth]{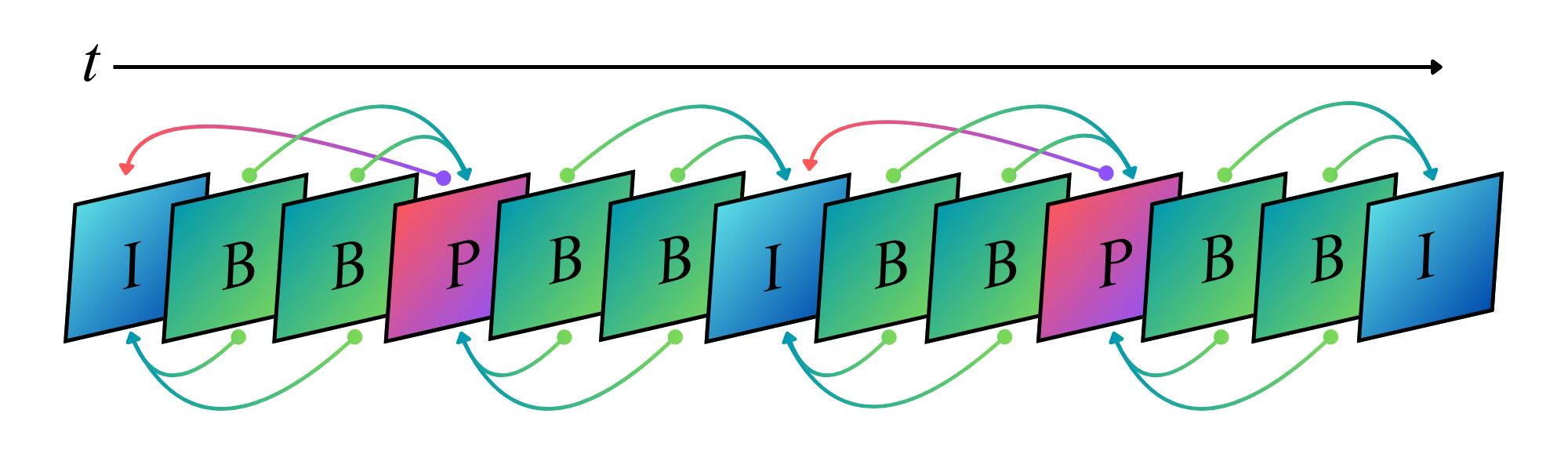}
    \caption{Group of Pictures (GOP) in H.264. Each group starts with an I-frame which is encoded independently. P and B frames are encoded based on their difference from the reference I or P frames. Due to this differential encoding, the packet-size variations are informative regarding the total change in the scene.}
    \label{fig:video_codec}
    \vspace{-0.15in}
\end{figure}

\subsection{Background}
\label{subsec:back_gop}

We focus on videos encoded using the H.264/MPEG-4 codec, one of the most widely used video compression standards \cite{wiegand2003overview}. H.264 enables efficient transmission by partitioning the stream into a \textbf{Group of Pictures (GOP)}, a set of consecutive frames encoded jointly. 
A visual description of GOP structure is given in Figure~\ref{fig:video_codec}. 
Typically, the first frame in each GOP, the \textbf{I-frame (Intra-coded)}, is encoded independently, while the remaining frames are encoded differentially to reduce temporal redundancy. \textbf{P-frames (Predictive)} are encoded using motion-compensated prediction from preceding reference frames (I- or P-frames).
B-frames (bidirectional frames) exploit both past and future reference frames for compression, but this bidirectional dependency inherently requires re-ordering and buffering, thereby increasing end-to-end latency. For latency-sensitive applications such as live video surveillance, many systems therefore disable or minimize the usage of B-frames in order to reduce delay~\cite{li2016side}. In this work, we utilize a configuration of I and P-frames without B-frames.
Because of this differential coding structure, the encrypted packet sizes convey information about the innovation, the change in the scene between consecutive frames.

\section{Problem Formulation}
We let $t$ denote the frame index and $T$ the total number of frames in an experiment. The position of the object at frame $t$ is represented by the vector $\mathbf{p}_t$. We define an object as \textbf{visible} at frame $t$ if it is in the Field of View (FoV) of at least one sensor in the network. 

The sensor network consists of $N_B$ \textbf{Blue Nodes} and $N_G$ \textbf{Gray Nodes}. 
\begin{itemize}
    \item \textbf{Blue Nodes:} Cameras with accessible raw signal-level video streams. The input from the $k$-th Blue Node is denoted as $\mathbf{X}^{(B)}_{k}$, representing a sequence of vectors of length $T$.
    \item \textbf{Gray Nodes:} Cameras where only sniffed packet-level bit-rates are accessible. The input from the $k$-th Gray Node is $\mathbf{X}^{(G)}_{k}$, which is also a sequence of length $T$.
\end{itemize}

Our objective is to perform inference on the observed scene, specifically tracking a vehicle's trajectory, using the set of packet size sequences $\{\mathbf{X}^{(G)}_{k}\}_{k=1}^{N_G}$, with or without the raw video data $\{\mathbf{X}^{(B)}_{k}\}_{k=1}^{N_B}$.

\subsection{System Assumptions and Constraints}
\label{sub_sec:assumption}
To ensure the tractability of the tracking problem and the consistency of the learned features, \ourSystem operates under the following assumptions:

\begin{enumerate}
    \item \textbf{Single Object and Background:} We assume a single-object constraint, where at most one target resides within the FoV of sensors at any time. Furthermore, the background is assumed to be largely static.
    
    \item \textbf{Codec and GoP Settings:} The video encoding parameters, specifically Group of Pictures (GoP) structure, are fixed. 
    
    \item \textbf{Sensor Configuration:} The sensor network has a fixed configuration of identical nodes. We assume both the intrinsic parameters and the poses of each node remain fixed between training and inference.
\end{enumerate}

The practical implications of these assumptions are further analyzed in Section~\ref{sec_dis}, where we also outline potential strategies to relax these constraints and extend the framework's versatility to more complex, dynamic environments.

\section{Solution Overview}\label{sec_system}

\begin{figure*}[!htp]
    \centering
    \includegraphics[width=0.9\linewidth]{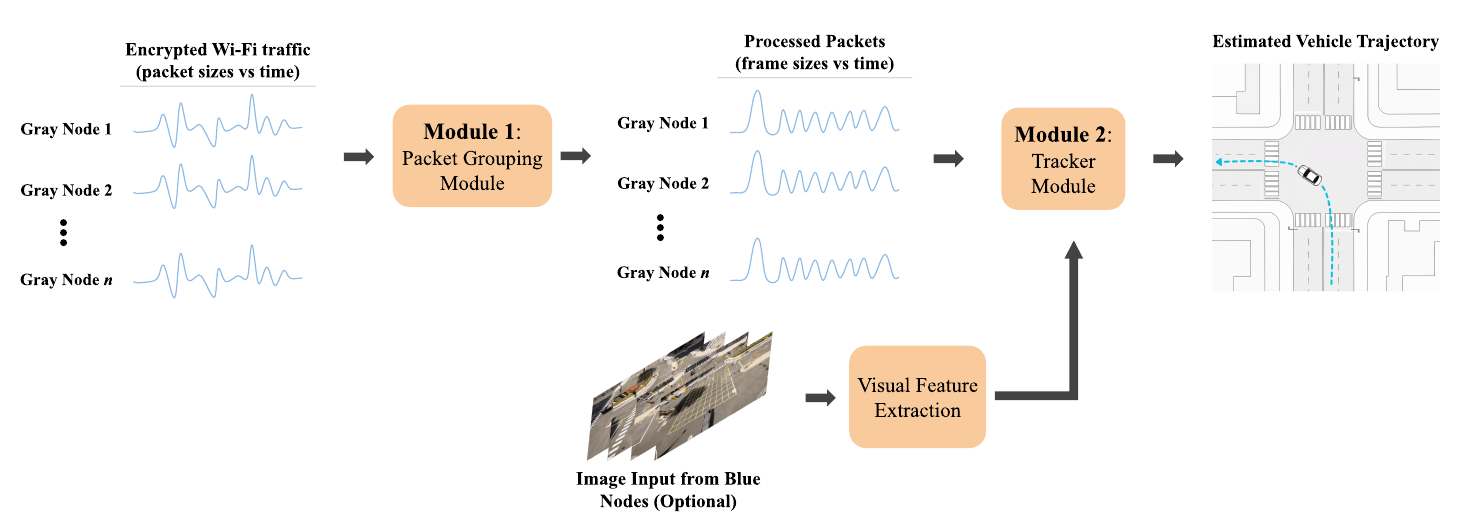}
    \caption{System overview of \ourSystem. Information on frame sizes extracted by \moduleOne along with the extracted image information is fed to the \moduleTwo module which estimates object's visibility by the sensors and its position when it is visible.}
    \label{fig:system_overview}
\end{figure*}

\ourSystem is a learning-based framework that estimates an object's trajectory by processing encrypted packet-size information from the \grayNodes. It can also fuse this indirect information with data from accessible camera streams, \blueNodes, when available.

As shown in Figure \ref{fig:system_overview}, the framework's first stage, the \moduleOne module, processes the raw stream of encrypted WiFi packets. Its primary function is to extract a structured time series of estimated frame sizes from the sniffed packets. This is a non-trivial task, as the observed packet information is often noisy due to network factors like bandwidth limitations or delay, limiting the effectiveness of inflexible time-windowing-based approaches~\cite{li2016side,mari2021looking} to packet trace segmentation. To overcome this, we use a transformer encoder-based model that learns to distinguish the boundaries of transmitted frames under network noise, allowing for accurate frame size reconstruction. Additionally, if a \blueNode is available, the raw images are processed by a Convolutional Neural Network (CNN) to extract useful visual features. 

The grouped packet-size time series and optional image features are then passed to the \moduleTwo module. Using a Transformer Encoder with a recurrent state to maintain consistency for tracking, \moduleTwo processes windows of the input data to perform two sequential tasks. It performs binary classification for vehicle's presence in the scene, followed by a position estimation if a vehicle is detected.

For clarity, we define several terms used throughout this work.
The \textit{packet size} refers to the raw size (in bytes) of each encrypted network packet captured from a video transmission stream - these are the inputs to the \moduleOne module. The \textit{grouped packet size} or \textit{extracted/estimated frame size} denotes the total size of all packets corresponding to an encoded video frame, estimated by the \moduleOne module. 
For comparison, the \textit{(raw) frame size} refers to the ground-truth frame-size information obtained directly from the original encoded video, prior to transmission over the network. We use extracted frame sizes generated by the \moduleOne module as inputs to the \moduleTwo module unless stated otherwise.

\section{Solution Details}

This section provides the details of our solution approach. We begin with a geometry-based analysis that explains how \textit{raw frame size} information enables object tracking. We then verify this approach experimentally in a controlled environment, using the results to build intuition and offer a possible explanation for the inner dynamics of our learning-based method. Finally, we describe the neural network architectures and loss functions for the packet size extraction and tracking stages in \ourSystem.

\subsection{Geometric Solution: From Frame Sizes to Position via Projection Area}

While H.264 frame sizes only provide an aggregate, non-spatial measure of the change in the scene, \ourSystem recovers spatial information by fusing frame size information from distinct views. This works because when the background is static, the change in the scene, hence the frame size, is dominated by object motion, and its magnitude is correlated with the object's projected area. This area is a function of the object's position, which allows \ourSystem to fuse measurements from non-degenerate views to localize the object.

\begin{figure}
    \centering
    \includegraphics[width=0.8\linewidth]{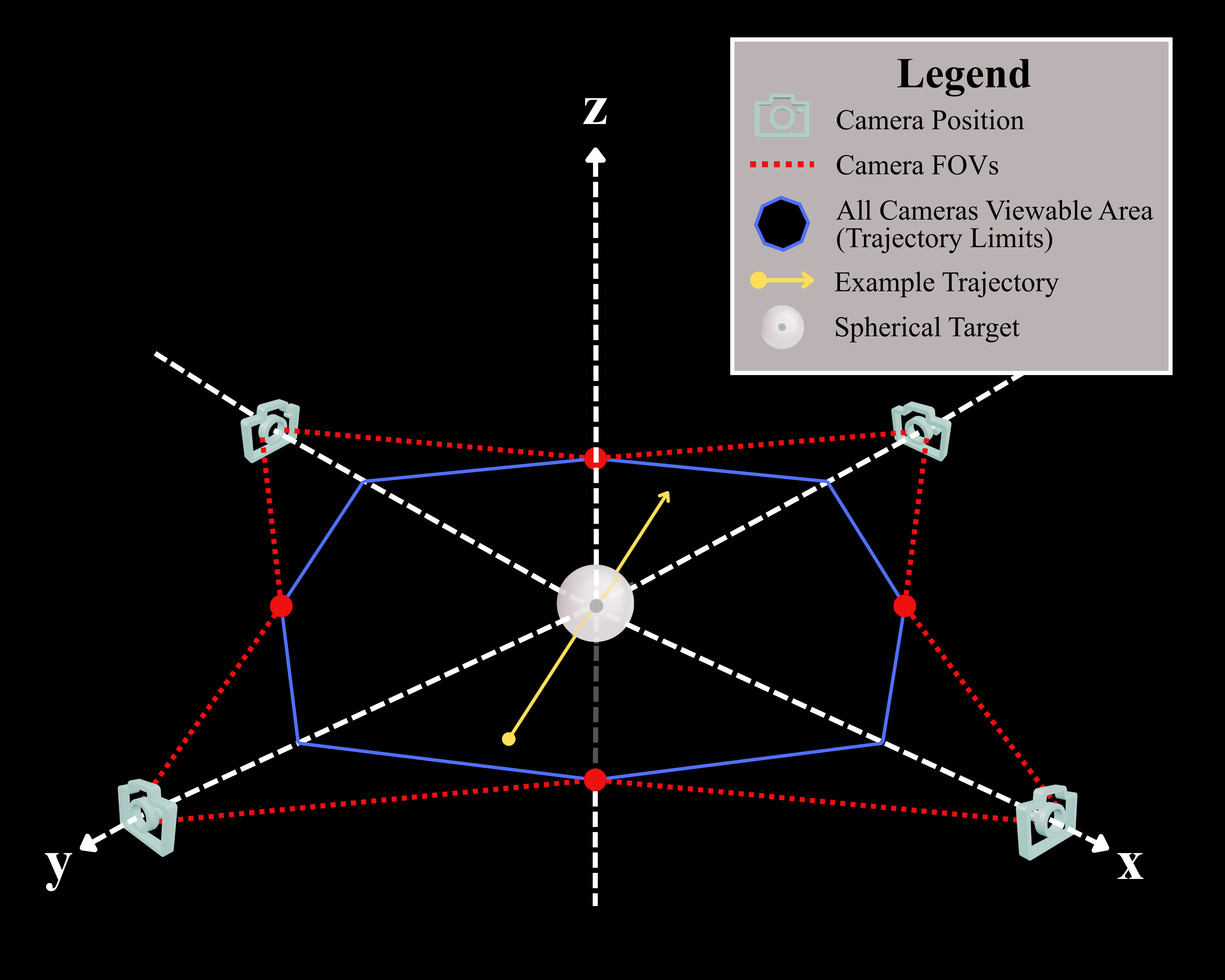}
    \caption{The Genesis setup used for geometric analysis. Four identical cameras are looking towards the sphere. The blue polygon shows the area in which the sphere moves, a region always visible to all cameras. In each experiment, the sphere moves at a random but constant velocity between randomly sampled initial and final points within this area.}
    \label{fig:genesis}
    \vspace{-0.15in}
\end{figure}

We hypothesize that the tracker module implicitly estimates this area for each \grayNode from the frame-size stream and, by fusing area estimates across distinct camera views, reconstructs the object’s position. We validate this hypothesis in a controlled environment using the Genesis simulator \cite{Genesis}. The setup is shown in Figure \ref{fig:genesis}.

To formalize the geometric problem, we define the sphere's center in world coordinates at time $t$ as $\mathbf{p}_t = [x_t, y_t, z]^\top$, where $z$ is known and fixed. For camera $i$ with extrinsic parameters (rotation matrix $\mathbf{R}_i$ and center $\mathbf{c}_i$), the sphere's center in the camera's coordinate system, $\mathbf{p}^{\mathrm{i}}_{t}$, is inferred by the transformation \cite{szeliski2010}:

\begin{equation}
    \mathbf{p}^{\mathrm{i}}_{t} = \mathbf{R}_i^\top(\mathbf{p}_t - \mathbf{c}_i)
    \label{eq:world_to_cam}
\end{equation}

\begin{figure}
    \centering
    \includegraphics[width=0.75\linewidth]{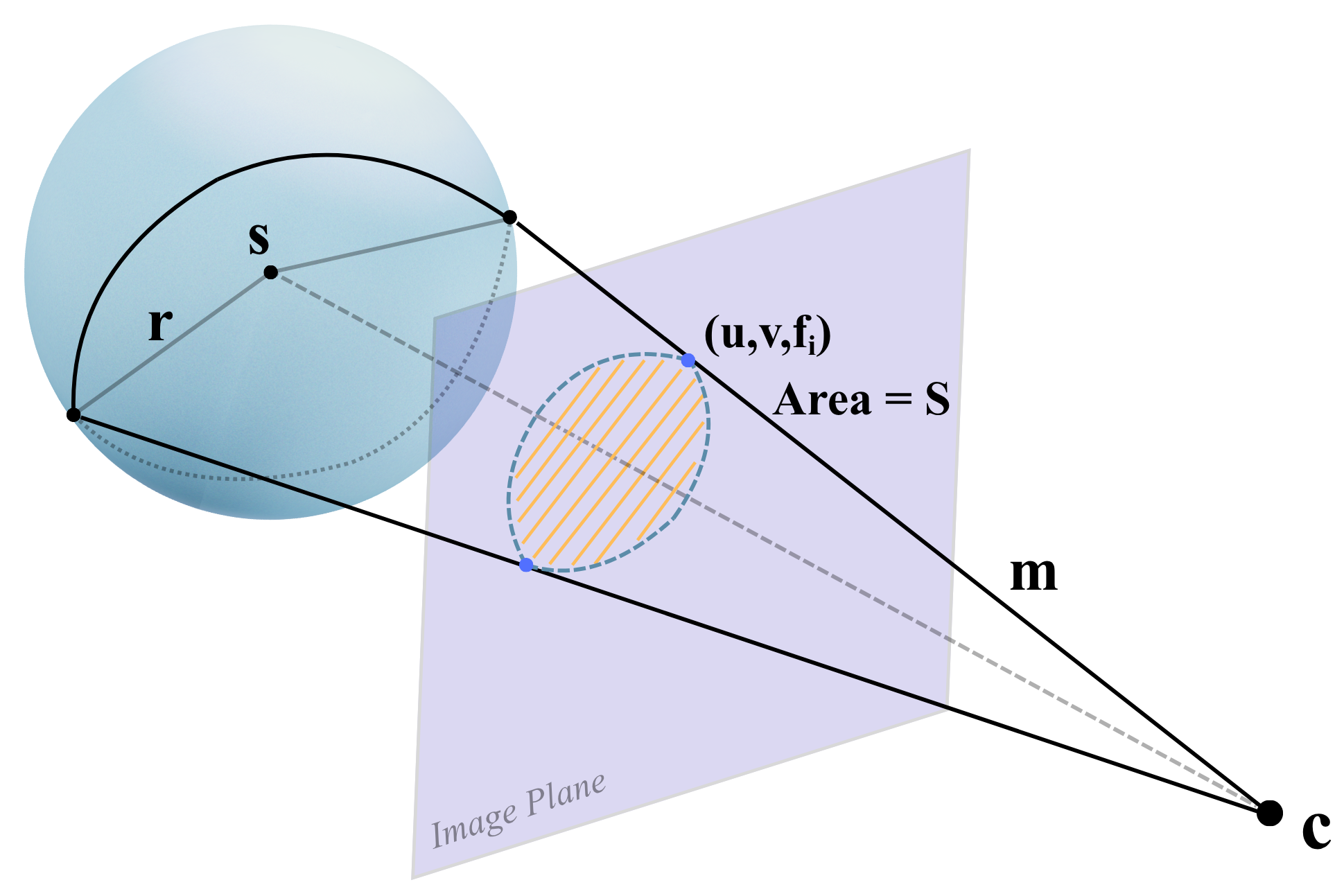}
    \caption{Geometric problem. The bold lines denote rays emanating from the camera center. Those rays that are tangent to the sphere intersect the image plane along an ellipse, forming the sphere’s silhouette in the image plane.}
    \label{fig:geometry} 
\end{figure}

To find the projected area, we first need to mathematically describe the sphere's silhouette on camera's image plane. A point $(u,v)$ on the image plane $(z = f_i)$ is part of the silhouette if the ray, $\mathbf{m}$ , originating from the camera center (origin) and passing through the point $[u, v, f_i]^\top$ is tangent to the sphere. This is visualized in Figure \ref{fig:geometry}.

The tangency condition requires the distance from the sphere's center $\mathbf{p}^{\mathrm{i}}_{t} = [x^i_t, y^i_t, z_t^i]^\top$ to the ray $\mathbf{m}$ to be equal to the sphere's radius $r$. Using the cross-product formula for the distance from a point to a line \cite{dunn2011}, we get the following:

\begin{equation}
    r = \frac{\|\mathbf{p_t^i} \times \alpha\mathbf{m}\|_2}{\|\alpha\mathbf{m}\|_2} = \frac{\|\mathbf{p_t^i} \times [u,v,f_i]^\top\|_2}{\|[u,v,f_i]^\top\|_2}.
    \label{eq:cross_product_distance}
\end{equation}

Simplifying Eq.~\eqref{eq:cross_product_distance} gives the quadratic equation for the silhouette (where $x_i, y_i, z_i$ are the coordinates of $\mathbf{p}^{\mathrm{i}}_{t}$):

\begin{equation}
    (x_i u + y_i v + z_i f_i)^2 - (u^2 + v^2 + f_i^2)(x_i^2 + y_i^2 + z_i^2 - r^2) = 0
    \label{eq:quadratic_silhouette}
\end{equation}

This equation describes an ellipse on the image plane with the center:

\begin{equation}
    (u_o,v_o) = \left(\frac{z_i f_i}{z_i^2-r^2}x_i, \frac{z_i f_i}{z_i^2-r^2}y_i\right)
    \label{eq:ellipse_center}
\end{equation}

For a centered ellipse in the form $Au^2 + Buv + Cv^2 = 1$, the area formula is given as \cite{korn2000}:

\begin{equation}
    S(A,B,C) = \frac{2\pi}{\sqrt{4AC-B^2}}
    \label{eq:area_ellipse}
\end{equation}

We translate the ellipse to the origin by substituting $u = u' + u_o$ and $v = v' + v_o$ to apply Eq. \eqref{eq:area_ellipse} and obtain the desired area expression for the sphere's projection onto the image plane:

\begin{equation}
S^i_{t} = \frac{\pi r^2 f_i^2}{\bigl((z_i)^2 - r^2\bigr)^{3/2}} \sqrt{(x_i)^2 + (y_i)^2 + (z_i)^2 - r^2}
\label{eq:area}
\end{equation}

Substituting Eq. \eqref{eq:world_to_cam} into Eq. \eqref{eq:area} shows that $S^{(i)}_{t}$ is a nonlinear function of the unknown world coordinates $(x_t,y_t)$. With $(\mathbf{R}^{(i)},\mathbf{c}^{(i)},f^{(i)})$ and $r$ known, area estimates from multiple cameras form a nonlinear system whose solution recovers $\mathbf{p}_t$.

We test our hypothesis by a two-stage procedure. First, we train an MLP to map true areas ${S^{i}_{t}}$ to position $\mathbf{p}_t$ (Eq.~\eqref{eq:area}). Second, we train a Transformer-based model, as in \ourSystem's \moduleTwo module, to estimate areas ${\hat S^{i}_{t}}$ from frame size inputs. Both modules perform well independently, as shown in Table \ref{tab:genesis}. 

We then freeze and connect them, feeding the Transformer’s estimated areas into the MLP, which attains a $0.43$~m mean position error, close to the $0.37$ m error of end-to-end \ourSystem on the same dataset, supporting the belief that \ourSystem implicitly learns projected area as an intermediate step for position estimation.

\begin{table}[!htbp]
\centering
\caption{Genesis simulation results.} 
\label{tab:genesis}
\setlength{\tabcolsep}{4pt} 
\begin{tabular}{cccc}
\toprule
\makecell{\textbf{Frame-to-Area} \\ \textbf{(Rel. Err.)}} & 
\makecell{\textbf{Area-to-Pos.} \\ \textbf{(m)}} & 
\makecell{\textbf{Pos.-from-Area} \\ \textbf{(m)}} & 
\makecell{\textbf{End-to-End} \\ \textbf{(m)}} \\
\midrule
$0.030 \pm 0.032$ & $0.123 \pm 0.041$ & $0.433 \pm 0.258$ & $0.368 \pm 0.217$ \\
\bottomrule
\end{tabular}
\end{table}

\subsection{\moduleOne Module: Extracting Frame Information from Encrypted Streams}

In our geometric analysis, we hypothesize the \textit{frame size} of an encoded frame carries information about changes in the camera's view, and hence can be used for tracking. However, in a realistic streaming setting, frames may be fragmented into many packets depending on the network configuration, and furthermore, when considering \grayNodes we assume only the encrypted packet streams are observable, so we lack the metadata to perfectly sort packets into their originating frames. To solve this, we use a two-stage approach, with an initial \moduleOne module that partitions the encrypted packet stream into groups corresponding to each frame, which are then passed to a \moduleTwo module. In this subsection, we focus on the design of the first stage.

In our experiments, we consider a network with bandwidth limit, delay, and jitter in the transmission, but we assume that the order of transmission remains unchanged. Consequently, we can pose the packet grouping as a boundary prediction problem: determining which packets signal the end of a frame, then summing the packet sizes between neighboring boundaries to construct a proxy for the frame sizes. 

The grouping module uses a transformer encoder architecture, with an input token of two features - the size of a sniffed packet in bytes ($b$) and the time difference between the current and previous packet transmit times ($\Delta t$). We use a fixed positional embedding to denote each token's place within the current window $n$, and an MLP head is used to output logits ($s_n$) for each packet in the window, whose sigmoid is used for final boundary probabilities. 

We train the model to minimize the sum of two loss terms, the binary cross-entropy (BCE) between true and predicted boundary labels, and a count loss that penalizes deviations between the true and predicted number of boundaries in each window:

\begin{equation}
\mathcal{L}^{\text{total}}_n = \
\mathcal{L}^{\text{boundary}}_n \;+\; \lambda_{\text{count}}\,\mathcal{L}^{\text{count}}_n.
\label{eq:stage1_l_total}
\end{equation}

The \textbf{Boundary Loss} (${L}^{\text{boundary}}_n$) rewards correct predictions of whether a packet represents the end of the current frame. The truth vector $y_n\in[0,1]$ is a boolean vector with 1's marking the last packet in each frame. The model outputs a boundary prediction logit for each packet in the window, and we compute the BCE between the sigmoid of the predicted logits and the true boundary label. Here we use the BCE between the sigmoid and ground truth over the 0/1 loss for its smoothness, which should result in faster training convergence.

\begin{equation}
\mathcal{L}^{\mathrm{boundary}}_n = \mathrm{BCE}(\sigma(\hat{s}_n), y_n).
\label{eq:stage1_l_boundary}
\end{equation}

The \textbf{Count Loss} term (${L}^{\text{count}}_n$) attempts to ensure the total number of predicted frames equals the true count in each window. This term is motivated by the need to keep the estimated frame size sequence temporally aligned with the ground-truth trajectory used in the \moduleTwo module. If the predicted and true number of frames differ across multiple windows, missing or extra frames can accumulate, causing a misalignment between the estimated frame indices and their corresponding ground-truth positions. To promote frame count consistency, we penalize the absolute difference between the true frame count and the sum of the boundary logits:

\begin{equation}
\mathcal{L}_n^{\text{count}}
= {\big|}\textstyle\sum{\sigma(\hat{s}_n)} - \textstyle\sum{y_n}{\big|}.
\label{eq:stage1_l_count}
\end{equation}

During training, we present the model with overlapping windows of the packet trace, with a window step size given by $\ell_{\text{stride}}$. The model generates a boundary prediction logit for each packet index in the window, then the losses are calculated and used to update the model using standard backpropagation. 

\begin{algorithm}
\caption{\moduleOne Training Procedure}
\label{alg:stage_1_training}
\begin{algorithmic}[1]
\Require Model $\theta$, $\lambda_{\text{count}}$, $n_{\text{stride}}$
\Require Training sequence $(\mathbf{X}, \mathbf{Y})$ (features $\mathbf{X}$, ground truth $\mathbf{Y}$)
\For{each window $n=1, 2, \dots, N_{\text{windows}}$}
    \State $(b_n,\Delta t_n)\leftarrow$ GetInputWindow($\mathbf{X}$, $n$)
    \State $y_n \leftarrow$ GetGroundTruth($\mathbf{Y}$, $n$)
    
    \State $\hat{s}_n\leftarrow \text{Model}(b_n,\Delta t_n; \theta)$ \Comment{Forward pass}
    
    \State $\triangleright$ \textit{Calculate window losses}
    \State $\mathcal{L}^{\mathrm{boundary}}_n \leftarrow \mathrm{BCE}(\sigma(\hat{s}_n), y_n)$
    \State $\mathcal{L}^{\mathrm{count}}_n \leftarrow {\big|}\textstyle\sum{\sigma(\hat{s}_n)} - \textstyle\sum{y_n}{\big|}$
    \State $\mathcal{L}^{\mathrm{total}}_n \leftarrow \mathcal{L}^{\text{boundary}}_n \;+\; \lambda_{\text{count}}\,\mathcal{L}^{\text{count}}_n$
    \State Update $\theta$ using $\nabla_\theta \mathcal{L}^{\mathrm{total}}$  \Comment{Perform update}
\EndFor
\end{algorithmic}
\end{algorithm}

\begin{algorithm}
\caption{\moduleOne Inference Procedure}
\label{alg:stage_1_inference}
\begin{algorithmic}[1]
\Require Model $\theta$, $\lambda_{\text{count}}$, $\ell_{\text{stride}}$, $\ell_{\text{window}}$, $\ell_{\text{max}}$
\Require Input packet trace $\mathbf{X}$
\State $\hat{y}^{(1 \times \ell_{\text{max}})} \leftarrow \{0\}^{(1 \times \ell_{\text{max}})}$ \Comment{Initialize output prediction buffer}
\State $z^{(1 \times \ell_{\text{max}})} \leftarrow \{0\}^{(1 \times \ell_{\text{max}})}$ \Comment{Initialize window counts buffer }

\For{each window $n=1, 2, \dots, N_{\text{windows}}$}
    \State $(b_n,\Delta t_n)\leftarrow$ GetInputWindow($\mathbf{X}$, $n$)
    
    \State $\hat{s}_n \leftarrow \text{Model}(b_n,\Delta t_n; \theta)$ \Comment{Forward Pass}

    \State $\triangleright$ \textit{Accumulate predictions and window counts}
    \State $\ell_{\text{start}} \leftarrow (n-1)\ell_{\text{stride}}$
    \State $\ell_{\text{end}} \leftarrow \ell_{\text{start}} + \ell_{\text{window}}$
    \State $\hat{y}_{[\ell_{\text{start}}:\ell_{\text{end}}]} \leftarrow \hat{y}_{[\ell_{\text{start}}:\ell_{\text{end}}]} + \sigma(\hat{s}_n)$
    \State $z_{[\ell_{\text{start}}:\ell_{\text{end}}]} \leftarrow z_{[\ell_{\text{start}}:\ell_{\text{end}}]} + \{1\}^{(1 \times \ell_{\text{window}})}$

\EndFor
\State $\hat{y} \leftarrow \hat{y} \, \oslash \, z$ \Comment{Average all window predictions}
\State $\hat{y} \leftarrow \mathbbm{1}[\hat{y}\geq0.5]$ \Comment{Round the averaged prediction}
\State \Return $\hat{y}$

\end{algorithmic}
\end{algorithm}

At inference time, we again present the model with overlapping windows with stride $\ell_{\text{stride}}$. For each window the model generates a boundary prediction ($\sigma(\hat{s}_n)$), which is added to an output prediction buffer ($\hat{y}$) at the indices corresponding to the window. A counter buffer ($z$) is also incremented, which records how many windows produced predictions for each packet index. After all window predictions have completed, the prediction buffer is divided element-wise by the counter buffer, to output the averaged predictions per packet index. Finally, those averaged predictions are rounded to $[0,1]$ by thresholding, yielding the final boundary predictions. The sums of the packets between each pair of boundaries (right-inclusive) are output as the reconstructed frame sizes.

The training and inference algorithms for the Stage 1 grouping model are given in Algorithms \ref{alg:stage_1_training} and \ref{alg:stage_1_inference}. We provide an analysis of the performance of our \moduleTwo module, and compare to a naive time-windowing-based grouping method in Table \ref{table:stage_1}.

\subsection{\moduleTwo Module: Estimating Positions from Frame Size Information}

\begin{figure*}[!ht]
    \centering
    \includegraphics[width=1\linewidth]{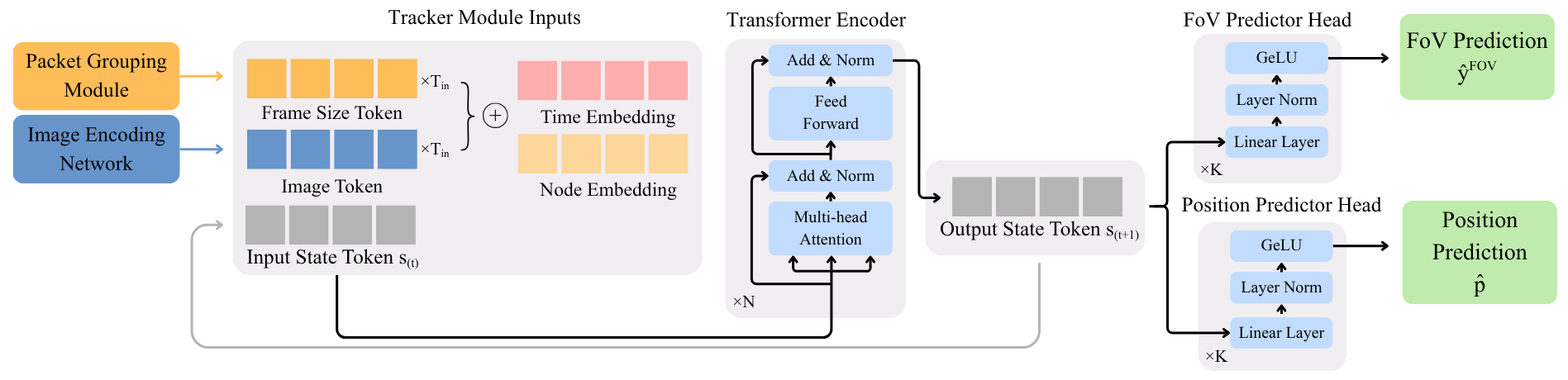}
        \caption{Tracker diagram with packet and image inputs. The extracted image and frame size vectors are tokenized and passed to the transformer along with the recurrent state token from the previous timestep. The output is retained as the next state token, and passed to the FoV and Position predictor heads.}
    \label{fig:tracker_networks}
    \vspace{-0.15in}
\end{figure*}

In this subsection, we describe the core of \ourSystem, the Tracker module.  As shown in Figure~\ref{fig:tracker_networks}, it takes a window of extracted frame sizes (grouped packet sizes) from \moduleOne module, along with images from the \blueNodes, if any, as input. The module's goal is to return the probability of an object's presence in the scene and a position estimation if the object is in the scene.

Given the sequential nature of the data, Transformer-based architectures are a natural choice for similar tasks \cite{carion2020end, meinhardt2022trackformer, zeng2022motr}. We employ a Transformer encoder equipped with a recurrent state token, a design similar to \cite{rm_transformer}. 

The module processes inputs within a sliding window of $T_{\mathrm{in}}$ frames. Grouped packet sizes come from the \moduleOne module for each \grayNode and \blueNode information is extracted by a Convolutional Neural Network (CNN). We then tokenize each element in the grouped packet size sequences along with the feature vectors for image frames. Before these sequences enter the Transformer, we augment each token by adding two learned embeddings: \textbf{Time embedding}, based on the token's position $i\in\{1,\dots,T_{\mathrm{in}}\}$ within the input sequence and \textbf{node embedding}, which identifies the token's source sensor $k\in\{1,\dots,N_I+N_D\}$.

For any given window $n$, the encoder processes the full sequence of packet and image tokens along with the state token from the previous window ($n-1$).

The resulting output state token serves a dual purpose. First, it is passed to two separate prediction heads to estimate the target's visibility ($\hat{y}_n^{\mathrm{fov}}$) and position ($\hat{\mathbf{p}}_n$). These heads are designed as two-layer MLPs with GeLU activation~\cite{hendrycks2016gaussian} and Layer Normalization~\cite{ba2016layer}. Second, the state token is carried over as the input state for the next window ($n+1$). This recurrent mechanism allows the model to maintain memory, extending its temporal context beyond $T_{\mathrm{in}}$ without increasing per-step computational complexity \cite{rm_transformer}. 

To explain the training and evaluation processes for the tracker, we first define some notation. $T_l$ is the total number of frames in experiment $l$. The input sequences are grouped packet size vectors of length $T_l$ from the \grayNodes ($\mathbf{X}^{(G)}_k$) and optionally $T_l$ frames from the \blueNodes ($\mathbf{X}^{(B)}_k$).

The tracker operates over a sliding window of $T_{\mathrm{in}}$ frames with a shift of $T_{\mathrm{stride}}$ frames between windows. For each window $n$, we define the target visibility probability $r_n$ and target position $\mathbf{p}_n$ by averaging over $T_{\mathrm{avg}}$ frames. Here, $r_n$ is the true fraction of frames the object is in the scene, and $\mathbf{p}_n$ is its average position over the frames it is visible. An object is declared visible in window $n$ if $r_n \ge \tau$, where $\tau$ is a required fraction of in-scene frames.

We train the model using a loss function with two components for each window $n$. The total loss is a weighted sum of the visibility loss and the position loss:
\begin{equation}
\mathcal{L}^{\mathrm{sum}}_n = \lambda_{\mathrm{fov}}\mathcal{L}^{\mathrm{fov}}_n + \lambda_{\mathrm{pos}} \mathcal{L}^{\mathrm{pos}}_n.
\label{eq:total_loss}
\end{equation}

\textbf{Visibility Loss ($\mathcal{L}^{\mathrm{fov}}_n$)} is for predicting the visibility of the object by the sensors. The ground truth $r_n \in [0,1]$ represents the fraction of frames the object is visible within the window $[t_n, t_n+T_{\mathrm{avg}}]$. The model outputs a corresponding prediction $\hat{y}_n^{\mathrm{fov}} \in [0,1]$, which we compare using the BCE loss. This loss is computed for every window as follows:
\begin{equation}
\mathcal{L}^{\mathrm{fov}}_n = \mathrm{BCE}(\hat{y}^{\mathrm{fov}}_n, r_n).
\label{eq:fov_loss}
\end{equation}

Second, the \textbf{Position Loss ($\mathcal{L}^{\mathrm{pos}}_n$)} is used conditionally. It is only computed if the object is deemed to be in the scene, defined by $r_n \geq \tau$. If this condition is met, the ground truth position $\mathbf{p}_n^{\mathrm{avg}}$ is the average of the object's true position over the frames it is visible in the window $[t_n, t_n+T_{\mathrm{avg}}]$. The position loss is the squared Euclidean distance:
\begin{equation}
\mathcal{L}^{\mathrm{pos}}_n = || \hat{\mathbf{p}}_n - \mathbf{p}_n^{\mathrm{avg}}||_2^2.
\label{eq:pos_loss}
\end{equation}
When $r_n < \tau$, the position loss is zero ($\mathcal{L}^{\mathrm{pos}}_n = 0$).

To train the recurrent model, we use Backpropagation Through Time (BPTT). We accumulate the loss over several windows:
\begin{equation}
\mathcal{L}^{\mathrm{cs}}_n = \mathcal{L}^{\mathrm{cs}}_{n-1} + \mathcal{L}^{\mathrm{sum}}_n.
\label{eq:cumulative_loss}
\end{equation}
The model parameters are updated every $f_{\mathrm{detach}}$ windows. After each update, the recurrent state token is detached from the computation graph, and the cumulative loss is reset to zero. This BPTT approach allows the model to learn a more stable and useful state representation. The complete training process is detailed in Algorithm~\ref{alg:Training}.

\begin{algorithm}
\caption{Tracker Training Process}
\label{alg:Training}
\begin{algorithmic}[1]
\Require Model $\theta$, $\lambda_{\mathrm{fov}}, \lambda_{\mathrm{pos}}, \tau, f_{\mathrm{detach}}$
\Require Training sequence $(\mathbf{X}, \mathbf{Y})$ (features $\mathbf{X}$, ground truth $\mathbf{Y}$)
\State Initialize recurrent state $s \leftarrow \mathbf{0}$
\State Initialize cumulative loss $\mathcal{L}^{\mathrm{cs}} \leftarrow 0$
\For{each window $n=1, 2, \dots, N_{\text{windows}}$}
    \State $W_n \leftarrow$ GetInputWindow($\mathbf{X}$, $n$)
    \State $(r_n, \mathbf{p}_n^{\mathrm{avg}}) \leftarrow$ GetGroundTruth($\mathbf{Y}$, $n$)
    
    \State $(\hat{y}_n^{\mathrm{fov}}, \hat{\mathbf{p}}_n, s_{\text{next}}) \leftarrow \text{Model}(W_n, s; \theta)$ \Comment{Forward pass}
    
    \State $\mathcal{L}^{\mathrm{fov}}_n \leftarrow \mathrm{BCE}(\hat{y}^{\mathrm{fov}}_n, r_n)$ \Comment{Calculate window losses}
    \If{$r_n \ge \tau$}
        \State $\mathcal{L}^{\mathrm{pos}}_n \leftarrow \| \hat{\mathbf{p}}_n - \mathbf{p}_n^{\mathrm{avg}}\|_2^2$
    \Else
        \State $\mathcal{L}^{\mathrm{pos}}_n \leftarrow 0$
    \EndIf
    \State $\mathcal{L}^{\mathrm{sum}}_n \leftarrow \lambda_{\mathrm{fov}}\mathcal{L}^{\mathrm{fov}}_n + \lambda_{\mathrm{pos}} \mathcal{L}^{\mathrm{pos}}_n$
    
    \State $\mathcal{L}^{\mathrm{cs}} \leftarrow \mathcal{L}^{\mathrm{cs}} + \mathcal{L}^{\mathrm{sum}}_n$ \Comment{Accumulate loss for BPTT}
    
    \State $\triangleright$ \textit{Perform periodic update and detach state}
    \If{$n \pmod{f_{\mathrm{detach}}} == 0$}
        \State Update $\theta$ using $\nabla_\theta \mathcal{L}^{\mathrm{cs}}$ 
        \State $\mathcal{L}^{\mathrm{cs}} \leftarrow 0$
        \State $s \leftarrow \text{detach}(s_{\text{next}})$ \Comment{Truncate BPTT graph}
    \Else
        \State $s \leftarrow s_{\text{next}}$
    \EndIf
\EndFor
\end{algorithmic}
\end{algorithm}

At inference time, the model operates sequentially. For each window $n$, the model processes the input (corresponding to frames in $[t_n, t_n+T_{\mathrm{in}}]$) to produce a visibility estimate $\hat{y}_n^{\mathrm{fov}}$ (for the interval $[t_n, t_n+T_{\mathrm{avg}}]$). For windows where $\hat{y}_n^{\mathrm{fov}} \geq \tau$, we also output the position estimate $\hat{\mathbf{p}}_n$. The model then advances to the next window, $t_{n+1} = t_n + T_{\mathrm{stride}}$, and repeats the process until the stream ends. As in training, the recurrent state token output from window $n$ is passed as the input state to window $n+1$, maintaining the model's temporal context. This inference procedure is summarized in Algorithm~\ref{alg:inference}.

\begin{algorithm}
\caption{Tracker Inference Procedure}
\label{alg:inference}
\begin{algorithmic}[1]
\Require Trained model parameters $\theta$, threshold $\tau$
\Require Input data stream $\mathbf{X}$
\State Initialize recurrent state $s \leftarrow \mathbf{0}$
\While{data stream $\mathbf{X}$ has windows}
    \State $n \leftarrow$ current window index
    \State $W_n \leftarrow$ GetInputWindow($\mathbf{X}$, $n$)
    \State $(\hat{y}_n^{\mathrm{fov}}, \hat{\mathbf{p}}_n, s_{\text{next}}) \leftarrow \text{Model}(W_n, s; \theta)$ \Comment{Forward pass}

    \If{$\hat{y}_n^{\mathrm{fov}} < \tau$}
        \State $\hat{\mathbf{p}}_n \leftarrow \varnothing$
    \EndIf
    \State \Return $\hat{\mathbf{y}}^{\mathrm{fov}}_n, \hat{\mathbf{p}}_n$
    \State $s \leftarrow s_{\text{next}}$ \Comment{Pass state to next window}
\EndWhile
\end{algorithmic}
\end{algorithm}

\section{Evaluation}\label{sec_eva}
A comprehensive evaluation was conducted to assess the performance, robustness, and generalizability of the \ourSystem system. The experiments were designed to validate the tracking performance and to systematically probe the system's resilience to a wide range of real-world conditions.

\subsection{Datasets}
\subsubsection{CARLA Simulation}
As shown in Figure~\ref{fig:carla_setup}, we generated a suite of synthetic datasets with realistic, high-fidelity vehicles, roads, and environments using the CARLA simulator~\cite{Dosovitskiy17}, which is a widely used tool used for urban driving simulation. Each dataset corresponds to a different driving scenario, designed to capture vehicle dynamics and diverse environmental conditions, e.g., vehicle types, colors, movements (trajectories and velocities), lighting (time of day), etc. 

We randomly spawn a vehicle from ["vehicle.tesla.model3", "vehicle.nissan.patrol", "vehicle.mercedes.sprinter"] which represent the 3 most common types (sedan, suv, and van) with each randomly given one of the 3 most common vehicle colors (white, black, and gray). We place four cameras in an intersection to monitor the scene at 30~fps. 
The vehicle’s movements can be controlled either by our custom scripts or by the BehaviorAgent. Basic maneuvers, such as going straight from the left lane in each of the four directions, are controlled by our scripts, which serve as the default setting. More complex behaviors, such as turning left, turning right, or going straight from both lanes, are handled by the BehaviorAgent provided by CARLA, as explored in Section~\ref{subsec:case_study}.
The vehicle’s initial or target velocity is randomly selected within the range of 5-20~m/s to emulate diverse driving dynamics, while the actual velocity may vary depending on the road geometry and the agent’s control behavior.
The CARLA simulator provides raw RGB frames, which we subsequently encode into H.264 videos~\cite{wiegand2003overview} using FFmpeg~\cite{ffmpeg} with the libx264 codec. 

\subsubsection{Network Emulation}

To emulate real-time transmission, the encoded videos are streamed via the Real-time Transport Protocol (RTP)~\cite{frederick2003rtp}.
Network conditions such as bandwidth (data rate) limits are simulated using Linux tc netem~\cite{hemminger2005network}, then delay and jitter are introduced to the packet timestamps, enabling fine-grained control over network variations. The network configurations used in our experiments are \emph{$[100,50,30,10]$} Mbps bandwidth limit, all with 20ms delay and 5ms random jitter, and their effects are explored in Section \ref{sec:main_results}

During streaming, we employ tcpdump and tshark~\cite{tcpdump,tshark} to capture the transmitted packets and extract packet-level infomation, including timestamps, packet lengths, \and ``last packet'' flags for each video frame. During training, the \moduleOne module uses the flags to learn to separate frames within the encrypted packet trace, while during the testing phase, only timestamps and packet lengths are provided. This pipeline enables the generation of realistic network-layer traffic traces with ground-truth video content and frame size for subsequent learning and analysis. 

\subsubsection{Ground-truth Noise}
\label{sec:datasets_gps_noise}
In a real-world deployment, \ourSystem would have to contend with fundamentally imperfect position data when learning to track targets. To obtain more grounded position data, we model the time-correlated error inherent in GPS measurements and add it to the precise position data from the CARLA simulator. We apply a first-order auto-regressive, AR(1), process \cite{10040237} independently to each horizontal coordinate, ignoring vertical error as vehicles in our experiments move in a plane.
\begin{equation}
e_{t+1} = \phi e_t + \eta_t, \quad \text{where } \phi = \exp{(-\Delta_t/\tau_{GPS})}.
\label{eq:gps_noise_rev}
\end{equation}
To capture the heavy-tailed nature of GPS errors in urban environments, the driving noise $\eta_t$ is drawn from a Student's t-distribution \cite{rs14225878, 6266757} with parameter $\nu$.

The process in Eq.~\eqref{eq:gps_noise_rev} is parameterized to be stationary. The driving noise variance, $\sigma_\eta^2 = \text{Var}(\eta_t)$, is set by $\sigma_\eta^2 = \sigma_e^2 (1 - \phi^2)$ to achieve a target steady-state, per-coordinate variance $\sigma_e^2$. The target total horizontal errors ($\sigma_H$) in our noise configurations were selected based on FAA WAAS performance standards \cite{FAA_WAAS_PAN_93}. Our experiments in Section \ref{sec:gps_noise} investigate the effect of GPS error on tracking performance under 4 noise settings: \emph{No GPS Noise}, \emph{Low Noise} 
($\sigma_H=0.64,\nu=9,\tau_{GPS}=60$), 
\emph{Medium Noise} 
($\sigma_H=1.00,\nu=5,\tau_{GPS}=300$),
and \emph{High Noise}
($\sigma_H=3.70,\nu=5,\tau_{GPS}=300$) \cite{Weaver2015GNSS}.

\begin{figure}[t]
    \centering
    \includegraphics[width=0.9\linewidth]{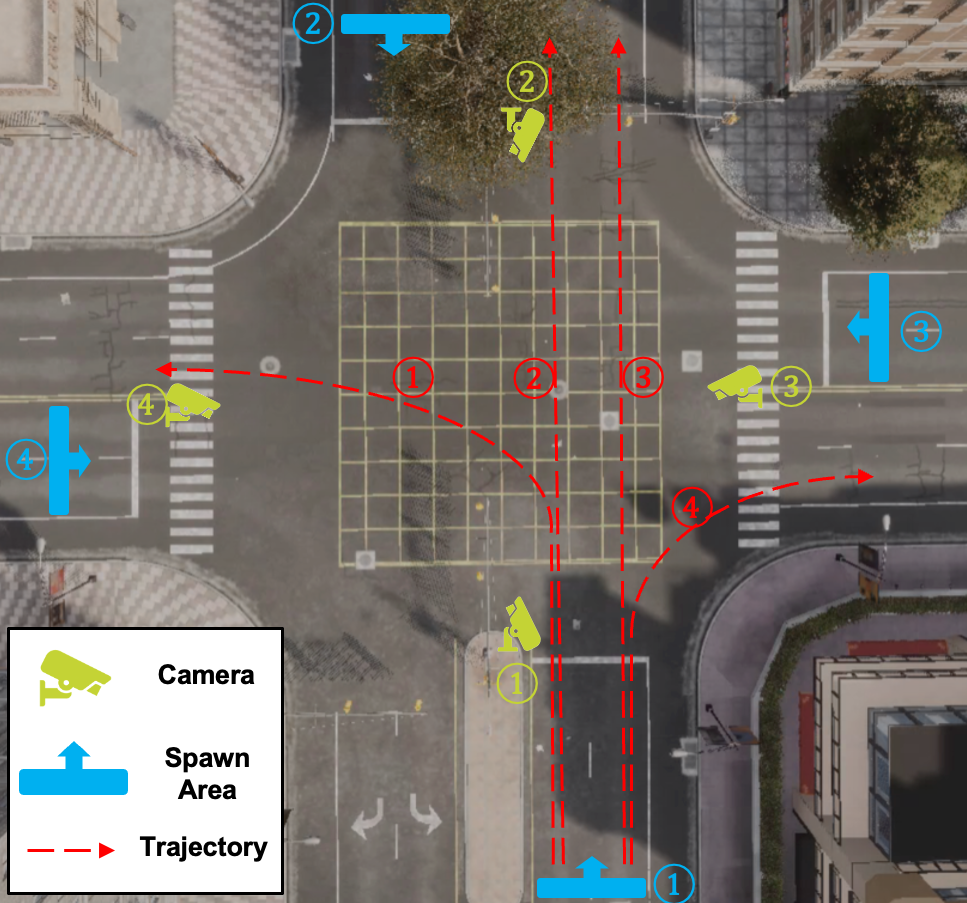}
    \caption{CARLA setup for synthetic data collection. 
    The environment consists of an intersection monitored by four cameras (green), which can serve as either \blueNodes{} or \grayNodes{} depending on the experiment configuration. 
    Vehicles are spawned from four starting areas (blue) and follow one of four possible trajectories or movements (red) \textit{per direction}.
    For clarity, only the four trajectories originating from Spawn Area~1 are illustrated.}
    \label{fig:carla_setup}
\end{figure}

\subsection{Implementation Details}
\label{subsec:implementation_details}

The datasets are generated on a desktop with an AMD Ryzen 9 9900X and an NVIDIA 5090 GPU. Model training and inference are performed on a server with an AMD EPYC~9354 and an NVIDIA H100 GPU, running Python~3.10.12, CUDA~11.8, and PyTorch~2.6.2. 

For the DNN used in stage 1, we use 4 transformer encoder layers with a token embedding size of $16$, a feed-forward dimension of $16$, and a single attention head. The stage 1 model is trained on a set of 1000 straight-trajectory samples independent of those used for training and evaluating stage 2, to avoid data leakage. For the DNN used in stage 2, we use 8 transformer encoder layers with a token embedding size of $128$, a feed-forward dimension of 512, and 4 attention heads, and the default parameters used are ($T_{\text{in}}=20,\,T_{\text{stride}}=10,\,T_{\text{avg}}=10,\,\tau=\frac{5}{6}$).  The parameter search to select the stage 2 default parameters is elaborated upon in Section~\ref{sec:stage2_param_search}. For experiments involving blue nodes, we rely on a pre-trained ResNet-18 as our backbone for extracting image features \cite{he2016deep}. To preserve the spatial information in the image features, we remove the pooling layer at the end of the pre-trained model and replace it with another convolutional layer with $128$ filters. 
This is the only part of the CNN model that is trained from scratch in our experiments involving blue nodes. 
The stage 2 train/test split is $80/20$ on a full dataset of 1000 vehicle trajectories per experiment, unless otherwise stated. We use the Adam optimizer with a learning rate of $2\times10^{-5}$ for the transformer layers, and $1\times10^{-4}$ for the MLP prediction heads, and a batch size 50. 
The default network configuration is $50$ Mbps bandwidth, $20$ ms delay, and $5$ ms jitter, and the default GPS noise level is set to \textit{Medium}, unless otherwise specified.

\subsection{Performance Metrics}
To evaluate performance, we use separate metrics for the two stages of our framework.
For Stage~1 (Packet Separation), we measure the boundary prediction error (the proportion of packets with boundary misclassifications) and the Dynamic Time Warping (DTW) distance between the predicted and ground-truth frame-size sequences, which captures the temporal alignment between series quality under jitter or delay.
For Stage~2 (Trajectory Estimation), we evaluate the field-of-view (FoV) prediction error, defined as the proportion of visible frames incorrectly classified as inside or outside at least 1 camera's FoV, and the L2 (Euclidean) distance between the predicted and ground-truth object locations, reflecting spatial tracking precision. All numerical results are reported on a test set, unseen by the model during training.

\begin{table*}[!htb]
\centering
\caption{Main results for reference upper bound settings and different network configurations}
\label{table:main_results}
\begin{tabular}{l l | c c c c}
\toprule
\textbf{Data} &&
\makecell{\textbf{Network Params}\\\textbf{(BW, Delay, Jitter)}} & 
\makecell{\textbf{Stage 1}\\\textbf{Pred. Error}} & 
\makecell{\textbf{Stage 2}\\\textbf{FoV Error (\%)}} & 
\makecell{\textbf{Stage 2}\\\textbf{Pos Error (m)}} \\
\midrule
Raw Video Access & (Highest Upper Bound)   &  - &  - &  
0.881 $\pm$ 1.266 &  0.780 $\pm$ 0.414 \\
Raw Frame Size & (Network Upper Bound)  &  - &  - &  2.669 $\pm$ 2.931 &  2.335 $\pm$ 3.487 \\
\midrule
\multirow{4}{*}{Packet size} 
&& 100Mbps, 20ms, 5ms &  (6.58 $\pm$ 8.10)$\,\times\,10^{-5}$  &  2.661 $\pm$ 3.150 &  2.961 $\pm$ 3.428 \\
&&  50Mbps, 20ms, 5ms &  (5.62 $\pm$ 7.92)$\,\times\,10^{-5}$  &  2.533 $\pm$ 2.812 &  2.334 $\pm$ 3.180 \\
&&  30Mbps, 20ms, 5ms &  (6.34 $\pm$ 8.79)$\,\times\,10^{-5}$  &  2.604 $\pm$ 2.861 &  2.597 $\pm$ 3.247 \\
&&  10Mbps, 20ms, 5ms &  (6.70 $\pm$ 9.42)$\,\times\,10^{-5}$  &  2.815 $\pm$ 3.056 &  2.907 $\pm$ 3.033 \\
\bottomrule
\end{tabular}
\end{table*}

\subsection{Overall Performance}
\label{sec:main_results}

We first evaluate the proposed framework under several representative settings, summarized in Table~\ref{table:main_results}. These configurations collectively characterize the performance gap between ideal access and the indirect, traffic-based sensing:
\begin{enumerate}
    \item \emph{Raw Video Access}: using direct camera streams to represent the best achievable performance with full raw image access.
    \item \emph{Raw Frame Size}: using the ground-truth frame-size sequence derived from encoded videos to reflect the best achievable results under network abstraction.
    \item \emph{Realistic Network Setting}: using only encrypted packet-level traffic captured under 100/50/30/10Mbps bandwidth (data rate) limit, 20 ms delay, and 5 ms jitter, representing typical real-world wireless video transmission conditions~\cite{vernersson2015analysis,chotalia2023performance}.
\end{enumerate}

Across all cases, our method demonstrates robust and consistent performance. Using raw images naturally yields the best results, as the system has direct access to pixel-level spatial information, representing the theoretical upper bound of performance. When only the ground-truth frame sizes are used (the network upper bound), the performance degrades due to the loss of explicit spatial structure, since frame sizes reflect only the aggregate magnitude of scene changes. Nevertheless, the resulting trajectory accuracy remains meaningful, especially considering that the smallest tracked vehicle in our dataset measures $4.61 \mathrm{m} \times 1.93 \mathrm{m}$.
Remarkably, when relying solely on encrypted packet-level traffic, \ourSystem achieves comparable tracking performance to the frame-size upper bound. It maintains a low Stage~1 error and achieves Stage~2 FoV and position errors at a similar scale to those obtained using ground-truth frame sizes. These results highlight the effectiveness of our two-stage design and demonstrate strong robustness to varying network conditions. Overall, the framework achieves reliable geospatial tracking using only encrypted traffic, confirming that network dynamics can encode sufficient information for meaningful motion inference.

\begin{table}[!htb]
\centering
\caption{Stage 1 Methods Comparison.}
\label{table:stage_1}
\begin{tabular}{ccccc}
\toprule
\makecell{\textbf{Grouping} \\ \textbf{Method}} &
\makecell{\textbf{Network Params} \\ \textbf{(BW, Delay, Jitter)}} &
\makecell{\textbf{Boundary} \\ \textbf{Pred. Err.} \\ \textbf{(1e-5)}} &
\makecell{\textbf{DTW Dist.} \\ \textbf{to True} \\ \textbf{Frame Sizes}} \\
\midrule
\multirow{4}{*}{\makecell{Learning\\-based \\(Ours)}}
 & 100Mbps, 20ms, 5ms & 6.58 $\pm$ 8.10 & 0.34 $\pm$ 1.48 \\
 &  50Mbps, 20ms, 5ms & 5.62 $\pm$ 7.92 & 0.32 $\pm$ 1.48 \\
 &  30Mbps, 20ms, 5ms & 6.34 $\pm$ 8.79 & 0.19 $\pm$ 0.92 \\
 &  10Mbps, 20ms, 5ms & 6.70 $\pm$ 9.42 & 0.16 $\pm$ 0.76 \\
\midrule
\multirow{4}{*}{\makecell{Time\\window}} 
 & 100Mbps, 20ms, 5ms & - & 9.83   $\pm$ 5.51  \\
 &  50Mbps, 20ms, 5ms & - & 70.87  $\pm$ 24.66 \\
 &  30Mbps, 20ms, 5ms & - & 107.34 $\pm$ 35.91 \\
 &  10Mbps, 20ms, 5ms & - & 291.33 $\pm$ 93.17 \\
\bottomrule
\end{tabular}
\end{table}

\subsubsection{Stage 1: Packet Grouping}

Table~\ref{table:stage_1} reports the Stage 1 results, comparing our learning-based extractor with a time-windowing baseline across different network conditions. 
This experiment evaluates the model’s robustness to varying bandwidth, delay, and jitter, which directly affect the temporal regularity of packet arrivals and thus the accuracy of frame-boundary detection. The baseline method applies a fixed-size time window to segment packets, a widely used preprocessing step in related work~\cite{li2016side,mari2021looking}, while our learning-based model learns temporal dependencies and compensates for irregularities introduced by network imperfections. The window size for the baseline is set to $1/30$~s, corresponding to the duration of one video frame at 30~fps.

As shown in Table~\ref{table:stage_1}, the time-window-based method performs well under high-bandwidth conditions (e.g., 100~Mbps) but degrades quickly as bandwidth decreases, with the DTW distance increasing substantially. This occurs because, at lower bandwidths, packets corresponding to large frames experience transmission delays and begin to overlap with or spill into the time slots of subsequent frames, making it difficult to correctly segment frame boundaries based on timing alone. In contrast, our learning-based extractor consistently achieves low boundary prediction errors and small DTW distances across all bandwidth settings. By learning robust temporal and structural features from packet sequences, it effectively separates frame boundaries even under severe bandwidth constraints, demonstrating strong resilience to network-induced temporal distortion.
We acknowledge the differences between our emulated network conditions and the non-stationary nature of real-world wireless links. These considerations are discussed in Section~\ref{sec_dis}.

\subsubsection{Stage 2: \moduleTwo Parameter Search}
\label{sec:stage2_param_search}

\newcommand\clearrow{\global\let\rowmac\relax}
\begin{table}[!htbp]
\centering
\caption{Parameter Search for Stage 2 DNN.}
\label{table:stage2_param_search}
\npdecimalsign{.}
\nprounddigits{3}
\begin{tabular}{cccccc}
\toprule
\textbf{$T_{\text{in}}$} & 
\textbf{$T_{\text{stride}}$} & 
\makecell{\textbf{($T_{\text{avg}}$, $\tau$)}} & 
\makecell{\textbf{State} \\ \textbf{Token} } & 
\makecell{\textbf{FoV} \\ \textbf{Error (\%)}} & 
\makecell{\textbf{Position} \\ \textbf{Error (m)}} \\
\midrule
20 & 10 & (10, 0.5) & w/  & 4.62 $\pm$ 3.34 & 4.05 $\pm$ 3.49 \\
10 & 10 & (10, 0.5) & w/  & 3.97 $\pm$ 3.12 & 3.80 $\pm$ 3.11 \\
40 & 10 & (10, 0.5) & w/  & 4.73 $\pm$ 3.31 & 4.76 $\pm$ 3.46 \\
20 & 5  & (10, 0.5) & w/  & 5.12 $\pm$ 2.75 & 5.39 $\pm$ 3.73 \\
\textbf{20} & \textbf{10} & \textbf{(6, 5/6)}  & \textbf{w/}  & \textbf{2.25  $\pm$ 2.70}  & \textbf{3.79  $\pm$ 3.19} \\
20 & 10 & (20,0.25) & w/  & 11.89 $\pm$ 4.91  & 5.79  $\pm$ 4.07 \\
20 & 10 & (10, 0.5) & w/o & 30.40 $\pm$ 9.00  & 25.04 $\pm$ 3.54 \\
\bottomrule
\end{tabular}
\end{table}

Table~\ref{table:stage2_param_search} presents an ablation study analyzing how the configuration of the Stage~2 DNN model influences performance under the default network configuration.
We vary the main hyper-parameters of the Stage~2 DNN, including the input window length ($T_{\text{in}}$), the stride between consecutive windows ($T_{\text{stride}}$), the averaging duration for generating ground-truth visibility and position labels ($T_{\text{avg}}$), and the FoV visibility threshold ($\tau$) that determines when a vehicle is considered within the FoV, along with the use of a recurrent state token.

Results show that incorporating a recurrent state token significantly improves the performance, reducing the average position error from $25.04$~m to $4.05$~m.  This is expected, as the state token allows the \moduleTwo to leverage information from previous windows for present FoV and position estimates, while marginally increasing the attention complexity. A moderate input length ($T_{\text{in}} = 20$) and stride ($T_{\text{stride}} = 10$) provide a good balance between accuracy and latency. 
In addition, a shorter averaging window ($T_{\text{avg}} = 6$) combined with a higher FoV threshold ($\tau = 5/6$) yields the most stable and accurate predictions, as it enforces stricter visibility confidence while preserving temporal smoothness. 
Based on these results, we adopt this configuration as the default setting for all other experiments.
These findings confirm that temporal aggregation with a recurrent memory effectively captures cross-frame dependencies in encrypted traffic sequences, enabling stable and precise trajectory estimation.

\subsection{Generalization and Robustness}
\label{subsec:case_study}

To further examine the generalization and robustness of our framework under diverse, real-world conditions, we conduct a series of experiments. These experiments explore how different scene dynamics and environmental factors affect \ourSystem's tracking performance.

\subsubsection{Complex Trajectories}
We evaluate the model’s ability to generalize to more complex motion patterns beyond basic straight-line trajectories. The CARLA BehaviorAgent controls the vehicle to execute various maneuvers, including turning left, turning right, and driving straight along two separate lanes, as shown in Figure~\ref{fig:carla_setup}. This increases the number of possible paths the vehicle may take from 4 to 16. When processing the raw packets for these experiments into framesizes, we refrain from re-training a \moduleOne model, instead reusing the weights learned using packet traces from straight trajectory experiments.

Our \moduleTwo module, when supplied directly with the frame sizes, achieves an FoV error of $3.211\,\pm\,2.783\%$, with an L2 position error of $1.863\,\pm\,2.098\text{m}$. When reconstructing frame sizes using the \moduleOne module, we observe a boundary prediction error of $(6.34\,\pm\,8.79)\times10^{-5}$, which is similar to the results we see when validating stage 1 on the same class of trajectories it was trained on. This illustrates that our stage 1 model learns to reconstruct frame sizes broadly, invariant to the specific content of the video streams. When using the reconstructed frame sizes produced from the \moduleOne module, under our standard network conditions, we observe an FoV error of $4.151\,\pm\,3.694\%$, with an L2 position error of $2.245\,\pm\,1.495\text{m}$. These results demonstrate that the learned temporal representation captures geometric relationships robustly across diverse trajectories and does not overfit to simple motions, both insulated and exposed to realistic network noise. 

\subsubsection{Ground-truth Position Noise}
\label{sec:gps_noise}
We introduce synthetic GPS noise modeled as an autoregressive AR(1) process with low, medium, and high noise levels to evaluate robustness to localization uncertainty. Our method of generating the synthetic noise is outlined in Section \ref{sec:datasets_gps_noise}.
As shown in Table~\ref{table:gps_noise}, both FoV and position errors increase slightly under low and medium noise conditions compared to the noise-free baseline, indicating graceful degradation under moderate noise. The errors rise further under high noise, which is expected since GPS noise directly affects the ground-truth positions used for training our framework, but at $5.13$m L2 position error, may still provide some useful tracking estimates.
These results demonstrate that the model can tolerate moderate inaccuracies in ground-truth trajectories - an important property given that real-world GPS annotations are often noisy. Moreover, higher-quality ground truth can always be obtained with more precise GPS devices or by integrating multiple positioning modalities, such as inertial localization, to further improve the performance of \ourSystem.

\begin{table}[h]
\centering
\caption{Impacts of GPS noise on tracking performance.}
\label{table:gps_noise}
\begin{tabular}{rcccc}
\toprule
\textbf{Noise Level} & \textbf{($\sigma_H,\nu,\tau_{GPS}$)} & \textbf{FoV Error (\%)} & \textbf{Pos Error (m)} \\
\midrule
None    & -               & 1.50 $\pm$ 2.20  &  1.86 $\pm$ 2.10 \\
Low     & (0.64,\,9,\,60) & 1.87 $\pm$ 2.51  &  2.36 $\pm$ 2.38 \\
Medium  & (1.00,\,5,\,300) & 2.53 $\pm$ 2.81  &  2.33 $\pm$ 3.18 \\
High    & (3.70,\,5,\,300) & 6.83 $\pm$ 4.70  &  5.13 $\pm$ 3.86 \\

\bottomrule
\end{tabular}

\end{table}

\subsubsection{Illumination}

To assess generalization across illumination conditions, we evaluate the tracking performance of the system throughout the day, at 10 time points from 8 AM to 5 PM, with 500 straight trajectory samples per period. The sun’s position is adjusted according to the geographic coordinates of Los Angeles on a typical October day to simulate realistic lighting changes throughout the day. 
As shown in Table~\ref{table:time_of_day}, both FoV and position errors remain stable across all illumination settings, with mean FoV errors between $2.82\%$ and $4.46\%$ and position errors ranging from $2.84$~m to $4.56$~m. These fluctuations in performance are minor relative to the overall scale of motion, demonstrating \ourSystem maintains consistent performance under substantial changes in lighting and shadow conditions.

\begin{table*}[!htb]
\small
\centering
\caption{Affect of time of day and lighting conditions on \ourSystem tracking performance.}
\label{table:time_of_day}
\setlength{\tabcolsep}{3pt}
\begin{tabular}{l l | *{10}{c}}
\toprule
\textbf{Training Set} & \textbf{Error Metric} & \textbf{08:00} & \textbf{09:00} & \textbf{10:00} & \textbf{11:00} & \textbf{12:00} & \textbf{13:00} & \textbf{14:00} & \textbf{15:00} & \textbf{16:00} & \textbf{17:00} \\
\midrule
\multirow{2}{*}{\makecell[l]{All times}} 
& \makecell{FoV Error (\%)} 
& 3.87 $\pm$ 3.68 & 4.05 $\pm$ 3.76 & 4.46 $\pm$ 4.17 & 3.60 $\pm$ 5.61 & 3.72 $\pm$ 3.60 & 3.31 $\pm$ 2.97 & 4.07 $\pm$ 4.32 & 3.57 $\pm$ 3.27 & 3.29 $\pm$ 3.46 & 2.82 $\pm$ 3.33 \\
& \makecell{Pos. Error (m)} 
& 3.44 $\pm$ 2.91 & 4.56 $\pm$ 5.65 & 3.52 $\pm$ 3.81 & 2.86 $\pm$ 2.13 & 3.10 $\pm$ 2.36 & 3.57 $\pm$ 3.85 & 2.84 $\pm$ 2.11 & 3.22 $\pm$ 3.39 & 2.99 $\pm$ 3.54 & 2.97 $\pm$ 3.55 \\

\bottomrule
\end{tabular}
\end{table*}

\subsection{Fusing Gray Nodes with a Blue Node}

\ourSystem can extend and complement the sensing capabilities of a \blueNode network by fusing its data with \grayNodes. We test this in the \textit{Complex Trajectories} setting by comparing two configurations: a single blue node setup and another one with three gray nodes in addition to the blue node. We focus on performance across two key scenarios: First when the object is visible only to \grayNodes and also the case where the object is visible to the \blueNode.

Results are presented in Table \ref{tab:blue_and_gray}. The most significant benefit of adding \grayNodes is the extension of the tracking range. In the scenario where the vehicle is visible only by the gray nodes, the \blueNode configuration fails, giving a very high FoV error (around $95\%$) and a very large position error of over $27$ m. This is expected scene the \blueNode cannot provide any meaningful information when the vehicle is not visible.  With the help of \grayNodes, \ourSystem improves these metrics by an order of magnitude, achieving an FoV error of $7.44 \%$ and a position error of $3.70$ meters. A slight, performance improvement is also observed when the object is visible to the blue node.

This extended sensing range can be observed in the weighted FoV error, average error over all windows, which decreases from $17.63\%$ to $2.65\%$. This demonstrates that \ourSystem successfully fuses information from both \blueNodes and \grayNodes, dramatically enhancing performance in new areas without decreasing the \blueNode's standalone capabilities.

\begin{table}[!htbp]
\centering
\caption{
    Comparison of single blue node and its fusion with three gray nodes. Average over the windows of $200$ validation experiments is reported.
}
\label{tab:blue_and_gray}
\setlength{\tabcolsep}{8pt} 
\begin{tabular}{l c c}
\toprule
\textbf{Error Metric} & \textbf{1 Blue} & \textbf{1 Blue + 3 Gray} \\
\midrule

\multicolumn{3}{l}{\textit{\textbf{Visible to Gray Nodes Only}}} \\
\quad FoV Error (\%)  & $95.11 \pm 15.15$ & $\textbf{7.44} \pm 18.35$ \\
\quad Pos. Error (m)  & $27.28 \pm 9.23$  & $\textbf{3.70} \pm 4.91$ \\
\midrule

\multicolumn{3}{l}{\textit{\textbf{Visible by the Blue Node}}} \\
\quad FoV Error (\%) & $2.92 \pm 3.84$ & $\textbf{0.90} \pm 2.39$ \\
\quad Pos. Error (m) & $1.86 \pm 0.95$ & $\textbf{1.44} \pm 0.726$ \\
\midrule

\multicolumn{3}{l}{\textit{\textbf{Summary}}} \\
\quad Weighted FoV Error & 17.63\% & \textbf{2.65\%} \\

\bottomrule
\end{tabular}
\end{table}
\vspace{-0.15in}
\section{Discussion and Future Work}\label{sec_dis}

While the proposed framework demonstrates that reliable tracking can be achieved using encrypted traffic, several open challenges and future directions remain to be explored.

\noindent\textbf{Tracking Multiple Objects.}
Our focus aligns with established paradigms in indirect sensing, such as WiFi-based human activity recognition and tracking \cite{li2016side, rasool2025invisible, ma2019wifi, abdelnasser2015ubibreathe}, which often concentrate on a single subject. Also, the single-object scenarios are of interest in settings like \textit{sterile zone} monitoring to detect and track intruders in high-stakes security perimeters \cite{nist_ir_7972, sandia_sttem} and in search and rescue scenarios \cite{uav_sar_tracking}.  
Although this work establishes the feasibility of encrypted traffic-based tracking, we recognize the inherent challenges of extending it to multi-object settings. 
Unlike traditional computer vision, where objects are spatially separated in the pixel domain, network traffic provides an aggregate signal of the entire scene. Therefore, multi-object tracking in this domain requires disentangling individual motion traces of different objects from a non-linear composition. This problem becomes more tractable when targets are spatio-temporally disjoint or exhibit distinct motion signatures. Furthermore, fusing these signals from \grayNodes with even a single \blueNode could serve as a powerful prior to resolve the ambiguities. We leave the formal solution to this problem for future work.

\noindent\textbf{Complex Motion and Behavior.}
Although our experiments include diverse trajectories such as turning, we have not explicitly tested behaviors like stopping at intersections. Modeling complex motions could enable applications in more realistic urban scenarios.

\noindent\textbf{Camera Calibration}
An interesting extension is to infer camera parameters automatically through calibration using controlled trajectories with GPS ground truth. For example, a vehicle equipped with accurate localization could traverse the scene to provide reference data for estimating camera positions, orientations, and FoV. While we have introduced small synthetic noise to mimic calibration errors, systematically analyzing their impact on performance remains an open problem.

\textbf{Codec Configuration Dependency}

While the current evaluation of \ourSystem utilizes the H.264 standard with a fixed Group of Pictures (GoP) structure, in practice such parameters can be inferred by analyzing packet-rate periodicity and temporal variations \cite{li2016side, rasool2025invisible}. Furthermore, the underlying principles of \ourSystem remain applicable to other codecs as contemporary standards such as H.265/HEVC \cite{sullivan2012overview}, H.266/VVC \cite{bross2021overview}, and AV1 \cite{han2020technical} all rely on similar inter-frame compression techniques to mitigate temporal redundancy. As these codecs consistently employ motion-compensated prediction, the resulting bit-rate fluctuations inherently mirror the physical dynamics of the scene. Consequently, \ourSystem is architecturally prepared to generalize across diverse video encoding profiles, provided that the specific GOP structure is known or inferred, allowing the model to match the structural characteristics of the specific compression profile in use.

\noindent\textbf{Real-world Deployment.}
All experiments in this work are conducted in simulation using CARLA and controlled network emulation, which provide a reproducible testbed. Translating this approach to real-world environments, where wireless interference, variable encryption protocols, and heterogeneous hardware coexist, will be crucial to validate scalability and robustness in practice.

\noindent\textbf{Online Inference.}
Currently, both stages operate in an offline, post-processing manner. Implementing the system in an online setup, processing packets as they arrive, would enable real-time tracking applications and integration with edge-based sensing platforms. 

\noindent\textbf{Privacy Risks and Mitigations.}
Like many sensing approaches, the proposed framework can potentially be misused to infer information that the data owner did not intend to reveal. Encrypted video traffic may leak side information about scene dynamics, raising privacy and security concerns if exploited by untrusted actors. Several mitigations exist to reduce such leakage, for instance, using constant-bitrate encoding, fixed packet sizes, or temporal padding to decouple bitrate from scene activity. Some of these defenses incur substantial costs in bandwidth, latency, and computation, or degrade visual quality and system responsiveness.
In practice, the ubiquity of legacy, resource-constrained cameras that cannot easily be updated makes such vulnerabilities persistent, offering an opportunity to exploit them to increase the sensory capabilities, as shown in our work. Acknowledging these trade-offs and studying how to balance them fairly across applications remains an important direction for future work.
Overall, these directions highlight the rich opportunities for advancing indirect perception through encrypted traffic. We believe this line of research can pave the way toward distributed sensing systems that reason over traffic dynamics as a complementary modality to visual data.
\section{Conclusion}\label{sec_conclusion}

In this work, we present \ourSystem, a learning-based framework that enables geospatial object tracking using encrypted video transmission traffic, without accessing visual content. By combining a \moduleOne module with a recurrent Transformer-based \moduleTwo, \ourSystem learns to infer frame structures and object motion directly from packet-level dynamics. Extensive experiments in CARLA simulations and emulated networks demonstrate that \ourSystem achieves reasonable tracking accuracy, revealing that encrypted traffic inherently encodes meaningful scene dynamics. 

\begin{acks}
The research reported in this paper was sponsored in part by the DEVCOM Army Research Laboratory under award \# W911NF1720196, and by the National Science Foundation under awards \# CNS-2211301, \#2502536. The views and conclusions contained in this document are those of the authors and should not be interpreted as representing the official policies, either expressed or implied, of the funding agencies. 
\end{acks}

\bibliographystyle{unsrt}
\bibliography{references}

\end{document}